\documentclass[pdflatex,sn-mathphys-num]{sn-jnl}

\usepackage{graphicx}
\usepackage{amsmath,amssymb,amsfonts}
\usepackage{xcolor}
\usepackage{textcomp}
\usepackage{manyfoot}
\DeclareNewFootnote{A}

\usepackage{booktabs}
\usepackage{url}

\begin{document}

\title[EnSol]{EnSol: an environment-aware graph neural network for molecular solubility prediction}

\author[1,2,4]{\fnm{Thao} \sur{Nguyen}}
\equalcont{These authors contributed equally to this work.}
\author[2,3,5]{\fnm{Saman} \sur{Shafaei}}
\equalcont{These authors contributed equally to this work.}
\author[2,3,5]{\fnm{Zhengyi} \sur{Zhang}}
\equalcont{These authors contributed equally to this work.}
\author*[2,3,4,5]{\fnm{Huimin} \sur{Zhao}}
\author*[1,2,4,5]{\fnm{Heng} \sur{Ji}}

\affil[1]{\orgdiv{Siebel School of Computing and Data Science}, \orgname{University of Illinois Urbana-Champaign}, \orgaddress{\city{Urbana}, \state{IL}, \postcode{61801}, \country{USA}}}
\affil[2]{\orgdiv{Carl R. Woese Institute for Genomic Biology}, \orgname{University of Illinois Urbana-Champaign}, \orgaddress{\city{Urbana}, \state{IL}, \postcode{61801}, \country{USA}}}
\affil[3]{\orgdiv{Department of Chemical and Biomolecular Engineering}, \orgname{University of Illinois Urbana-Champaign}, \orgaddress{\city{Urbana}, \state{IL}, \postcode{61801}, \country{USA}}}
\affil[4]{\orgdiv{NSF Molecule Maker Lab Institute}, \orgname{University of Illinois Urbana-Champaign}, \orgaddress{\city{Urbana}, \state{IL}, \postcode{61801}, \country{USA}}}
\affil[5]{\orgdiv{DOE Center for Advanced Bioenergy and Bioproducts Innovation}, \orgname{University of Illinois Urbana-Champaign}, \orgaddress{\city{Urbana}, \state{IL}, \postcode{61801}, \country{USA}}}

\miscnote{\textsuperscript{*}To whom correspondence should be addressed. Tel: (217) 244-0862, Email: hengji@illinois.edu (H.J.); (217) 333-2631, Email: zhao5@illinois.edu (H.Z.).}

\abstract{Molecular solubility directly affects key aspects of molecular development such as reaction feasibility, formulation performance, separation efficiency, and solvent selection. However, experimental measurement across solutes, solvents, and temperatures remains costly and sparsely sampled. Existing computational models often rely on fixed-solvent assumptions, deterministic formulations, or simplified representations of solute--solvent interactions, limiting their ability to capture complex molecular interactions, continuous temperature effects, and experimental uncertainty. Here, we introduce EnSol, an environment-aware probabilistic framework for molecular solubility prediction. EnSol represents the solute and solvent as molecular graphs and learns separate representations for each before bringing them together through cross-attention to capture solute--solvent interactions. Temperature is incorporated directly into the solvent environment through feature-wise modulation, and a mixture density network predicts full solubility distributions to capture both temperature-dependent behavior and experimental uncertainty. On the independent SolProp and Leeds benchmark datasets, EnSol achieved Spearman correlations of 0.876 and 0.601, respectively, outperforming state-of-the-art solubility prediction models across both benchmarks. Beyond computational benchmarking, experimental validation across chemically diverse solute--solvent pairs showed that EnSol maintained strong predictive performance and supported reliable solvent ranking, achieving a Spearman correlation of 0.715. These results show that EnSol can support reliable solubility prediction and solvent selection across diverse chemical systems while accounting for predictive uncertainty.}


\maketitle

\section{Main}\label{sec:main}

Molecular solubility is an important physicochemical property that influences the behavior and utility of small molecules across different solvents and environments.\cite{bib1,bib2,bib3} In practice, insufficient or poorly characterized molecular solubility constrains formulation stability, mass transfer efficiency, and reaction performance, often increasing experimental iteration and development timelines.\cite{bib4,bib5,bib6} In industrial and laboratory settings, suboptimal solubility can necessitate additional burden such as phase management, and downstream processing steps, increasing both operational complexity and cost.\cite{bib7,bib8} Despite its key role, systematic experimental characterization of molecular solubility across diverse solvent systems and temperature conditions remains labor-intensive, time-consuming, and costly, making large-scale screening difficult during early molecular development and process design.\cite{bib9,bib10,bib11} Consequently, reliable predictive models for molecular solubility offer a practical way to accelerate decision-making, reduce experimental bottlenecks, and support rational solvent and process selection.\cite{bib12,bib13}

Recently, machine learning (ML) has been increasingly applied to predict molecular solubility and prioritize solvents before extensive laboratory testing.\cite{bib14}, \cite{bib15},\cite{bib16} However, existing models often oversimplify this problem in ways that limit their ability for realistic solvent selection. Many approaches focus on aqueous solubility or fixed-solvent settings, whereas practical chemical workflows require comparison across many solvent environments.\cite{bib3},\cite{bib17} Other models incorporate solvent information but treat solute and solvent representations as static features, limiting their ability to learn specific interaction patterns among solvent and solutes.\cite{bib18},\cite{bib19} Furthermore, despite the direct influence of temperature on molecular dissolution and solvent behavior, it is often treated as a simple input variable rather than as a continuous environmental factor that modulates the solvation environment and solute--solvent interactions. Lastly, most solubility predictors return a single deterministic point as prediction, even though experimental measurements can vary across protocols, datasets, and thermodynamic conditions. As a result, these models provide little information about prediction uncertainty, making it harder to judge which predictions are reliable enough to guide experiments.\cite{bib6},\cite{bib20}

To address these limitations, we developed EnSol, an environment-aware probabilistic ML-based model for molecular solubility prediction. EnSol is designed around the idea that accurate solubility prediction requires modeling the full experimental context, including the solute, solvent, and temperature. The model represents solute and solvent molecules as molecular graphs and encodes them using graph neural network (GNN). Their learned representations are then coupled through cross-attention, allowing the model to construct interaction-aware features that depend on the specific solute--solvent pair. Temperature is introduced through modulation of the solvent representation, so changes in temperature can directly influence how the solvent environment is represented by the model. EnSol also moves beyond a single-point prediction to model a full distribution of possible solubility values, allowing the model to both account for heterogeneous or multimodal behavior and provide uncertainty estimates that help distinguish more confident predictions from those that may require additional experimental testing.

EnSol was designed to reflect the experimental setting as closely as possible by explicitly modeling the solubility task. We evaluate EnSol through two separate computational benchmarks and an independent experimental solvent-screening validation. Across all settings, EnSol consistently outperformed the comparison models, supporting its use for practical solubility prediction and solvent selection, showing how the model can move beyond retrospective prediction toward practical use in solvent screening by directly incorporating molecular context and environmental conditions into data-driven molecular solubility prediction.

\section{Results}\label{sec:results}

\subsection{Overview of the EnSol architecture}\label{sec:overview}

\begin{figure}[htbp]
\centering
\includegraphics[width=\textwidth,trim=0bp 1500bp 20bp 0bp,clip]{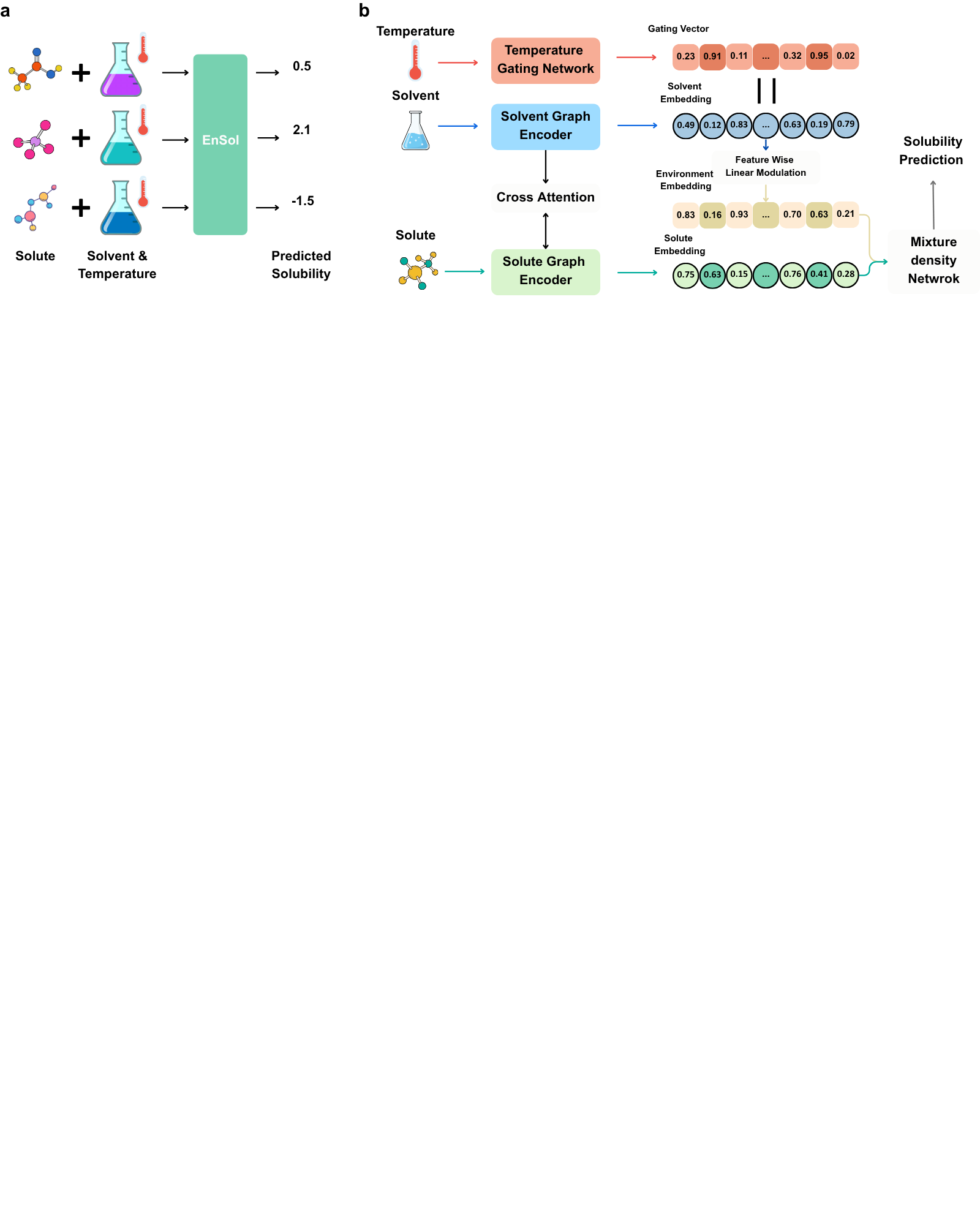}
\caption{\textbf{a}\textbf{)} Conceptual illustration of the solubility landscape, showing how a single solute exhibits condition-dependent solubility across different solvents and temperatures, motivating environment-aware modeling. \textbf{b}\textbf{)} Overview of the EnSol architecture. Solute and solvent molecules are encoded as graphs, coupled via cross-attention, and conditioned on temperature to form an environment embedding. A mixture density network predicts a full solubility distribution, enabling uncertainty-aware predictions.}\label{fig:1}
\end{figure}

EnSol is an environment-aware molecular solubility prediction framework that jointly models solute structure, solvent structure, and temperature within a unified deep learning architecture (\textbf{Fig.~\ref{fig:1}}). Solute and solvent molecules are represented as molecular graphs and encoded using GNN modules, after which their node-level representations are coupled through cross-attention to learn context-dependent solute--solvent interaction features. Temperature is incorporated as a continuous environmental variable that modulates the solvent embedding through feature-wise linear modulation (FiLM)\cite{bib21}, producing a unified solute--solvent--temperature representation of the experimental condition. This joint representation parameterizes a Mixture Density Network\textit{ (}MDN)\cite{bib22}, letting EnSol to predict full conditional solubility distributions rather than a point estimates. With integrating molecular interaction, continuous temperature conditioning, and probabilistic inference, EnSol captures heterogeneous solubility behavior across solvent and temperature regimes and provides uncertainty estimates for downstream screening and decision-making.

\subsection{Training and comparative evaluation of EnSol}\label{sec:benchmark}

EnSol was trained on BigSolDB\cite{bib23}, one of the largest and most complete databases of experimentally measured molecular solubility values across diverse organic solvents and temperatures (\textbf{Fig. }\textbf{S}\textbf{1}). Solute-level splitting was used for training and validation to avoid leakage, and performance was evaluated on two independent external benchmark datasets, SolProp\cite{bib24}, and Leeds\cite{bib6}, after removing solutes overlapping with the training set (\textbf{Fig. }\textbf{S}\textbf{2 and Fig. }\textbf{S}\textbf{3}).

\begin{figure}[htbp]
\centering
\includegraphics[width=\textwidth,trim=0bp 1300bp 70bp 0bp,clip]{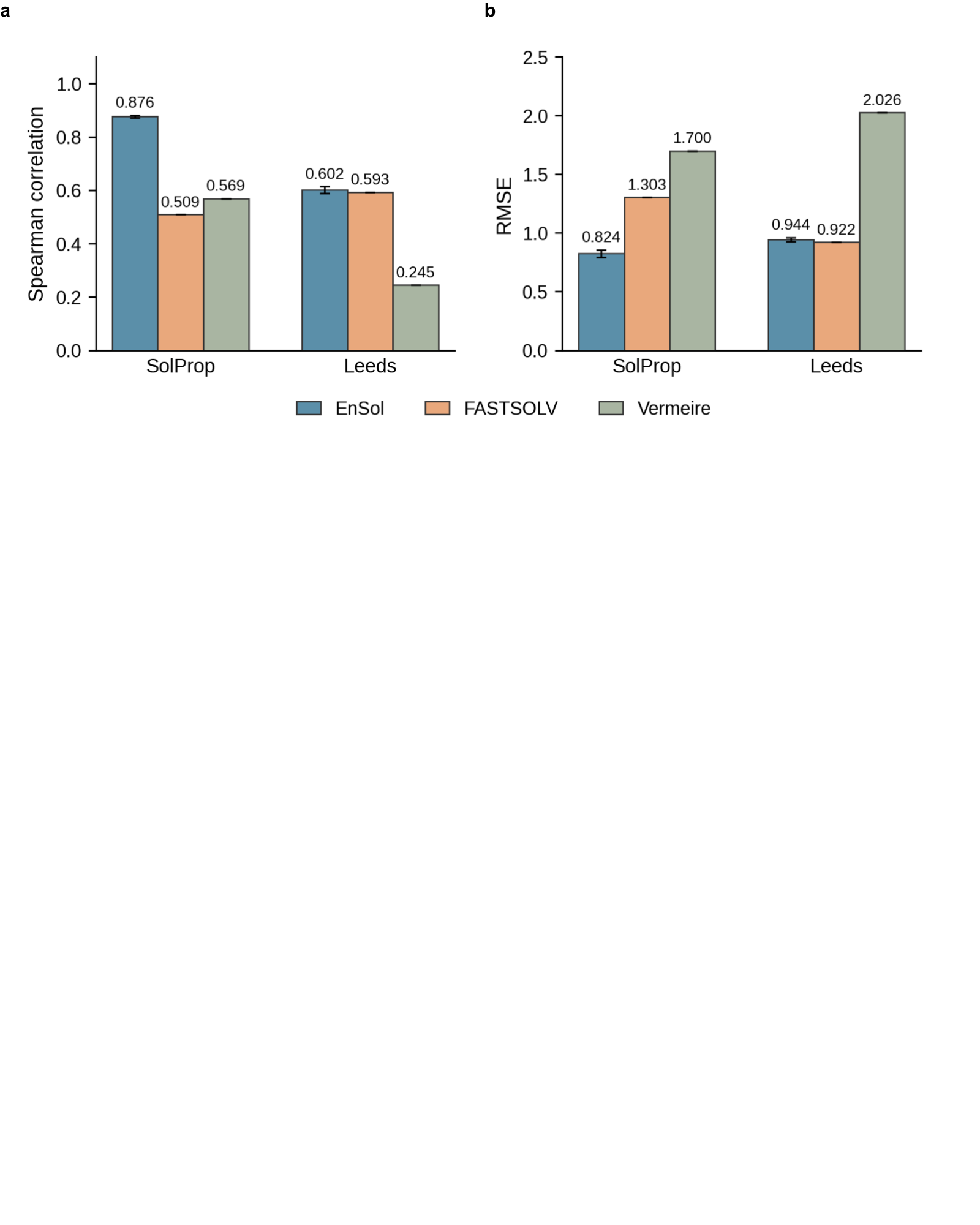}
\caption{Predictive accuracy and ranking performance across two independent benchmark datasets SolProp and Leeds. Model performance is compared on two external test sets, SolProp and Leeds. Values above bars are mean values. EnSol error bars are standard deviation across three random seeds (n = 3) and FASTSOLV and the Vermeire model are pretrained and deterministic. \textbf{a}\textbf{)} Spearman rank correlation \textbf{b}\textbf{)} RMSE values.}\label{fig:2}
\end{figure}

We benchmarked EnSol against two state-of-the-art solubility prediction models, FASTSOLV\cite{bib19} and Vermeire\cite{bib24}, using identical filtered test sets. EnSol performance is across three random seeds, whereas both baseline models are pretrained and deterministic. On the SolProp benchmark, EnSol showed a substantial improvement in solubility ranking, achieving a Spearman correlation of 0.876 $\pm$ 0.005 compared with 0.509 for FASTSOLV and 0.569 for the Vermeire model (\textbf{Fig.~\ref{fig:2}a}), while reducing RMSE to 0.824 $\pm$ 0.032 from 1.303 and 1.700, respectively (\textbf{Fig.~\ref{fig:2}}\textbf{b}\textbf{, }\textbf{Table} \textbf{S}\textbf{1 and }\textbf{Fig. S4a}).

On the Leeds benchmark, which is more distant from the training distribution, EnSol clearly outperformed the Vermeire model while closely matching FASTSOLV. EnSol achieved a Spearman correlation of 0.602 $\pm$ 0.013, compared with 0.593 for FASTSOLV and 0.245 for the Vermeire model (\textbf{Fig.~\ref{fig:2}a}). A similar pattern was observed for prediction error, as EnSol achieved a Root Mean Square Error (RMSE) of 0.944 $\pm$ 0.017, which was comparable to 0.922 for FASTSOLV and substantially lower than 2.026 for the Vermeire model (\textbf{Fig.~\ref{fig:2}}\textbf{b}\textbf{, }\textbf{Table} \textbf{S1} \textbf{and }\textbf{Fig. S4b}).

Because practical solvent selection often requires ranking candidate solvents for a fixed solute under defined experimental conditions, we further evaluated EnSol in a solvent-ranking setting using the SolProp database.\cite{bib25} Candidate solvents were prioritized for each solute--temperature pair according to the predicted solubility. Compared to FASTSOLV\cite{bib19}, EnSol identified high-solubility solvents among the top-ranked candidates more consistently and better maintained the relative ordering of solvent performance, reaching top 1 recall of 0.61 and top 3 recall of 0.85 versus 0.35 and 0.56 respectively (\textbf{Fig. }\textbf{S}\textbf{5}). This suggests EnSol can support solvent screening as a decision-making tool, rather than only serving as a point predictor of individual solubility measurements.

\subsection{Transferability and expert model fine-tuning}\label{sec:transfer}

Solubility models are often adapted to settings in which the relevant chemical space of solutes or solvent are narrow and only a limited number of measurements are available, such as a new compound series or a single solvent of interest. We therefore examined whether the representations learned by EnSol from the chemically diverse, multi-solvent BigSolDB dataset could transfer to aqueous solubility prediction, and whether the value of this pretraining correlates with the number and composition of the target data. To this objective, transfer learning capability of EnSol was evaluated on two complementary aqueous datasets. We first incorporated AqSolDB\cite{bib26}, a large curated collection of experimentally measured aqueous solubilities. After removing compounds that overlapped with the aqueous subset of BigSolDB, 9,748 unique compounds remained. We also included ESOL\cite{bib12}, a widely used aqueous solubility dataset containing 1,082 compounds. For each dataset, we trained EnSol in two ways. For each dataset, we compared fine-tuning EnSol from a BigSolDB-pretrained checkpoint with training the same architecture from random initialization using identical Murcko-scaffold splits and optimization settings across five seeds. Because the target datasets do not report temperature, both approaches used a temperature-independent version of EnSol. We additionally evaluated fixed subsets of 1,000 AqSolDB compounds to examine how transfer changed with target data availability.

\begin{figure}[htbp]
\centering
\includegraphics[width=\textwidth,trim=0bp 1310bp 125bp 0bp,clip]{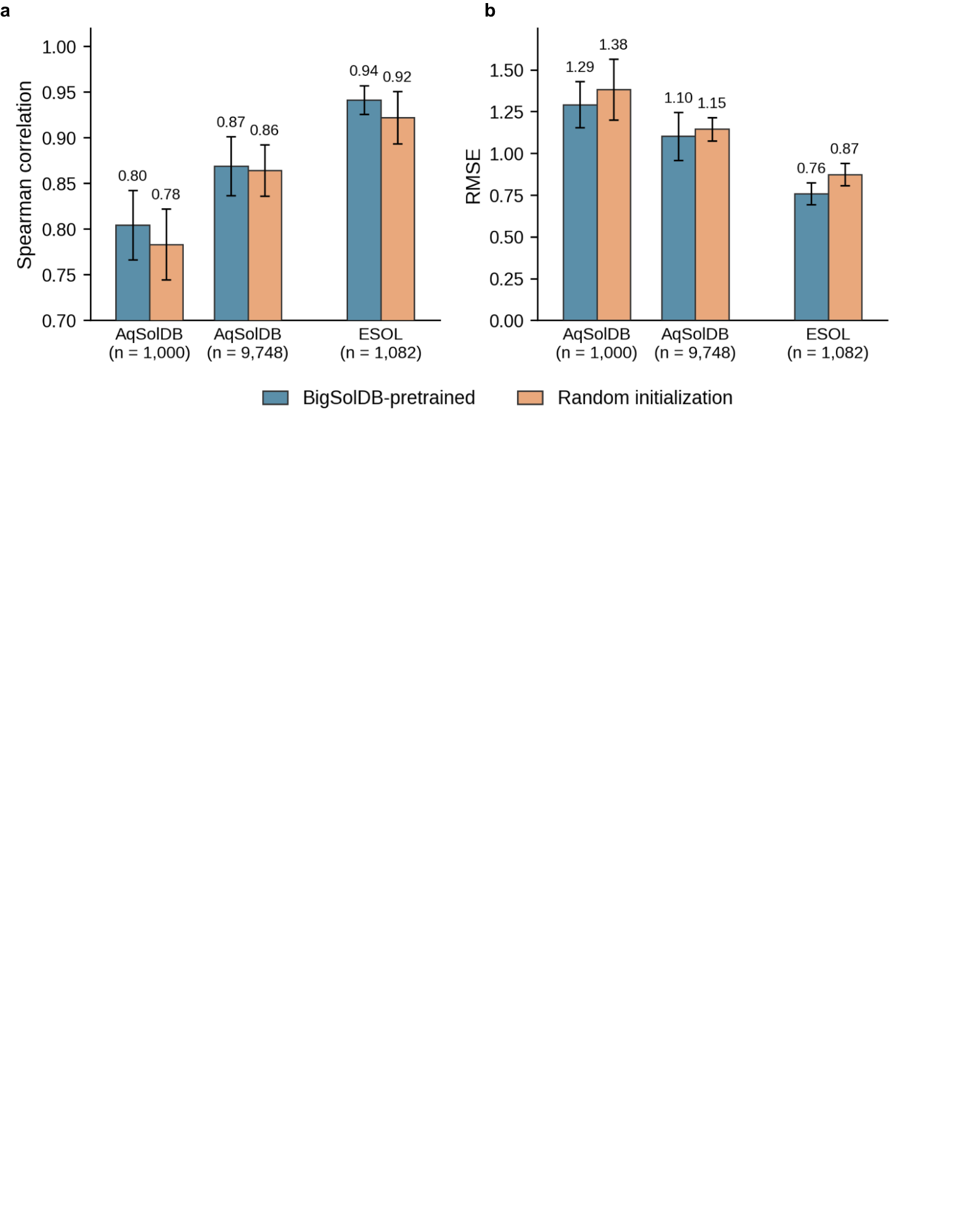}
\caption{Transfer learning and generalization of EnSol. Fine-tuning from a BigSolDB-pretrained checkpoint compared with training the identical architecture from random initialization, across three random seeds (n = 3) of the two deduplicated AqSolDB set (n = 1,000 and n= 9,748) probe the effect of target dataset size and ESOL (n = 1,082) is an independent aqueous dataset shown for comparison. Values above bars are means. a) Spearman rank correlation b) RMSE values.}\label{fig:3}
\end{figure}

Across both datasets, the benefit of pretraining was more apparent in RMSE than in Spearman correlation, suggesting that pretraining mainly helped EnSol predict solubility values more accurately, while having a more modest effect on the relative ranking of compounds. On ESOL, pretraining increased the Spearman correlation from 0.922 $\pm$ 0.029 to 0.942 $\pm$ 0.016 (\textbf{Fig.~\ref{fig:3}a}) while it had larger effect on the RMSE and reduced the RMSE from 0.874 $\pm$ 0.066 to 0.759 $\pm$ 0.065 (\textbf{Fig.~\ref{fig:3}b}). The benefit was also followed the same pattern in the data-limited AqSolDB setting. With 1,000 fine-tuning compounds, pretraining reduced the RMSE from 1.383 $\pm$ 0.181 to 1.292 $\pm$ 0.139 and increased the Spearman correlation from 0.783 $\pm$ 0.039 to 0.804 $\pm$ 0.038. Also, as more target-domain data were introduced, the contribution of pretraining gradually shrank. On the full AqSolDB dataset, pretraining produced only a small reduction in RMSE, from 1.146 $\pm$ 0.070 to 1.102 $\pm$ 0.143, while the Spearman correlation changed little, from 0.864 $\pm$ 0.028 to 0.869 $\pm$ 0.032 (\textbf{Table S2}). None of these differences reached statistically significant, indicating that the observed gains were modest relative to variation arising from initialization and scaffold partitioning and relationship between target dataset size and transfer benefit was not strictly monotonic.

The contrast between ESOL and AqSolDB further indicates that transfer depends on the composition and internal consistency of the target dataset rather than on sample size alone. Despite containing approximately nine-fold fewer compounds, ESOL supported substantially lower prediction error and higher rank correlation than the full AqSolDB dataset. This difference is consistent with the narrower and more uniformly curated chemical space of ESOL, whereas AqSolDB aggregates measurements from multiple sources and covers a broader and more heterogeneous distribution. Overall, the results indicate that BigSolDB pretraining provides a slight transferable initialization for aqueous solubility prediction, with the clearest gains observed when target measurements are limited. The benefit is more consistent for reducing absolute prediction error than for improving compound ranking and becomes less pronounced as sufficient task-specific data become available.

\subsection{Ablation studies and model component contributions}\label{sec:ablation}

To quantify the contribution of each individual architectural component within EnSol, we performed a systematic ablation study on SolProp test set, in which key modules of the model were selectively removed or replaced while keeping the rest of the parameters including training splits, optimization settings, and evaluation protocol fixed over three different seeds. This comparison allowed us to isolate the functional contribution of molecular encoding, solute--solvent interaction modeling via cross-attention, temperature conditioning, and probabilistic prediction to overall model performance.

\begin{figure}[htbp]
\centering
\includegraphics[width=\textwidth,trim=0bp 1325bp 45bp 0bp,clip]{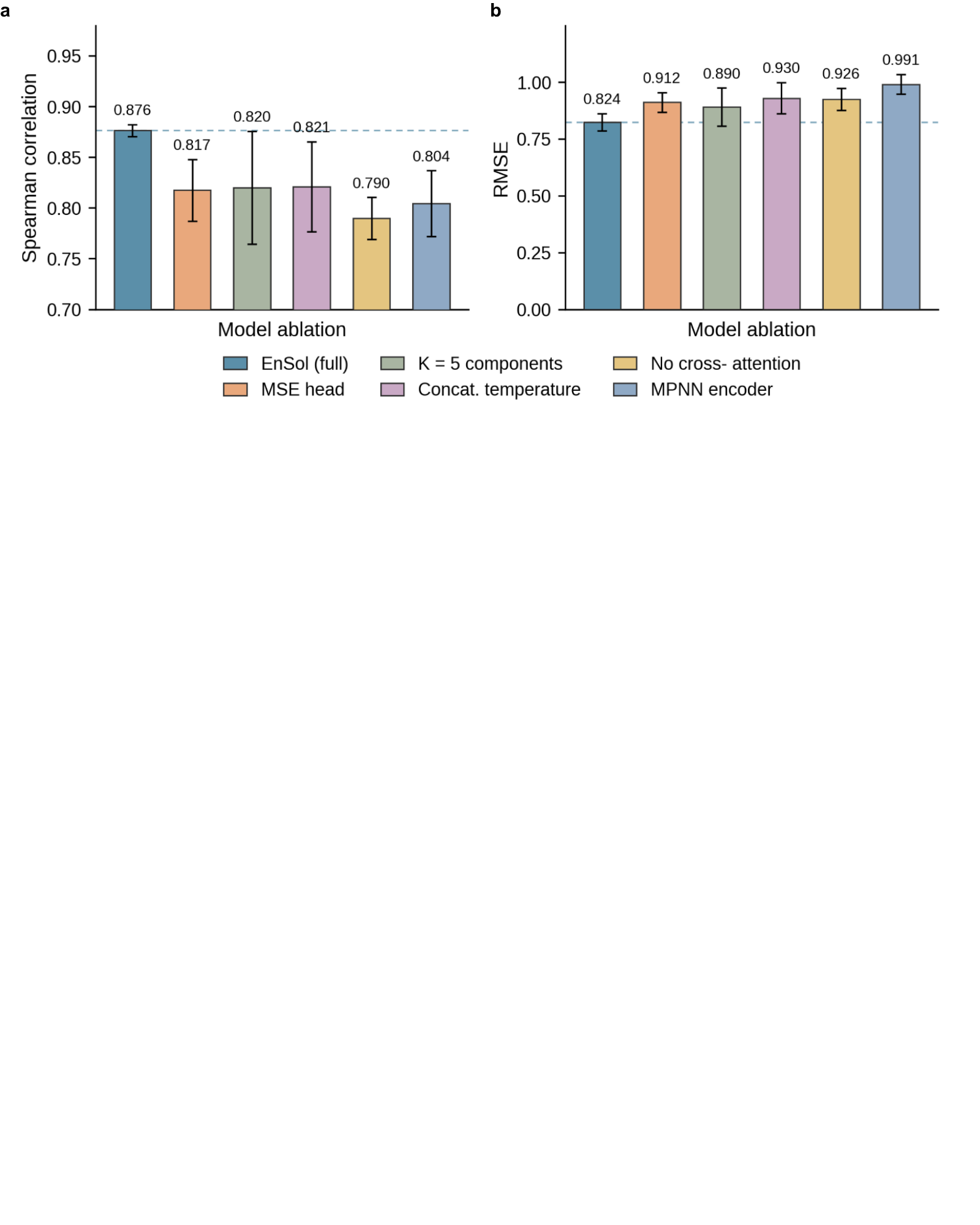}
\caption{Component-wise ablation of the EnSol architecture on the SolProp benchmark. Each variant modifies exactly one design choice relative to full EnSol, holding training splits, optimization settings, and evaluation protocol fixed. K = 5 components, the Gaussian mixture is increased from three to five components; No cross-attention, the solute--solvent cross-attention layer is removed; MSE head, the mixture-density head is replaced by a deterministic regression head trained with mean-squared error; Concat. temperature, FiLM temperature conditioning is replaced by a normalized temperature scalar concatenated onto the solute--solvent embedding; MPNN encoder, the AttentiveFP graph encoder is replaced by a message-passing neural network. Performance is shown as a) Spearman rank correlation. b) RMSE values.}\label{fig:4}
\end{figure}

The results indicated that molecular encoder and solute--solvent interaction mechanism made large contributions to performance. Replacing AttentiveFP\cite{bib26} with a conventional MPNN\cite{bib27}, corresponded to an average decrease of 0.088 in Spearman correlation and an increase of 0.187 in RMSE, which showcases the importance of the way the solvents and solute molecules are represented. Removing solute--solvent cross-attention also decreased Spearman correlation by 0.088 and increased RMSE by 0.104, showing that explicitly modeling interactions between the two molecular representations improves performance. Replacing FiLM-based temperature conditioning with direct temperature concatenation decreased Spearman correlation by 0.042 and increased RMSE by 0.092, further supporting our hypothesis of temperature-dependent modulation of the solvent representation. Similarly, replacing the mixture-density network with a deterministic mean squared error (MSE) regression head decreased Spearman correlation by 0.043 and increased RMSE by 0.100 (\textbf{Fig.~\ref{fig:4}a}\textbf{, }\textbf{Fig.~\ref{fig:4}b}\textbf{ and Table S3}). The combination of the ablation study results suggest that each major component contributes to EnSol performance, with the molecular encoder and solute--solvent cross-attention producing the largest improvements in ranking performance.

\subsection{Experimental validation and solvent recommendation}\label{sec:experimental-validation}

\begin{figure}[htbp]
\centering
\includegraphics[width=\textwidth,trim=0bp 995bp 0bp 0bp,clip]{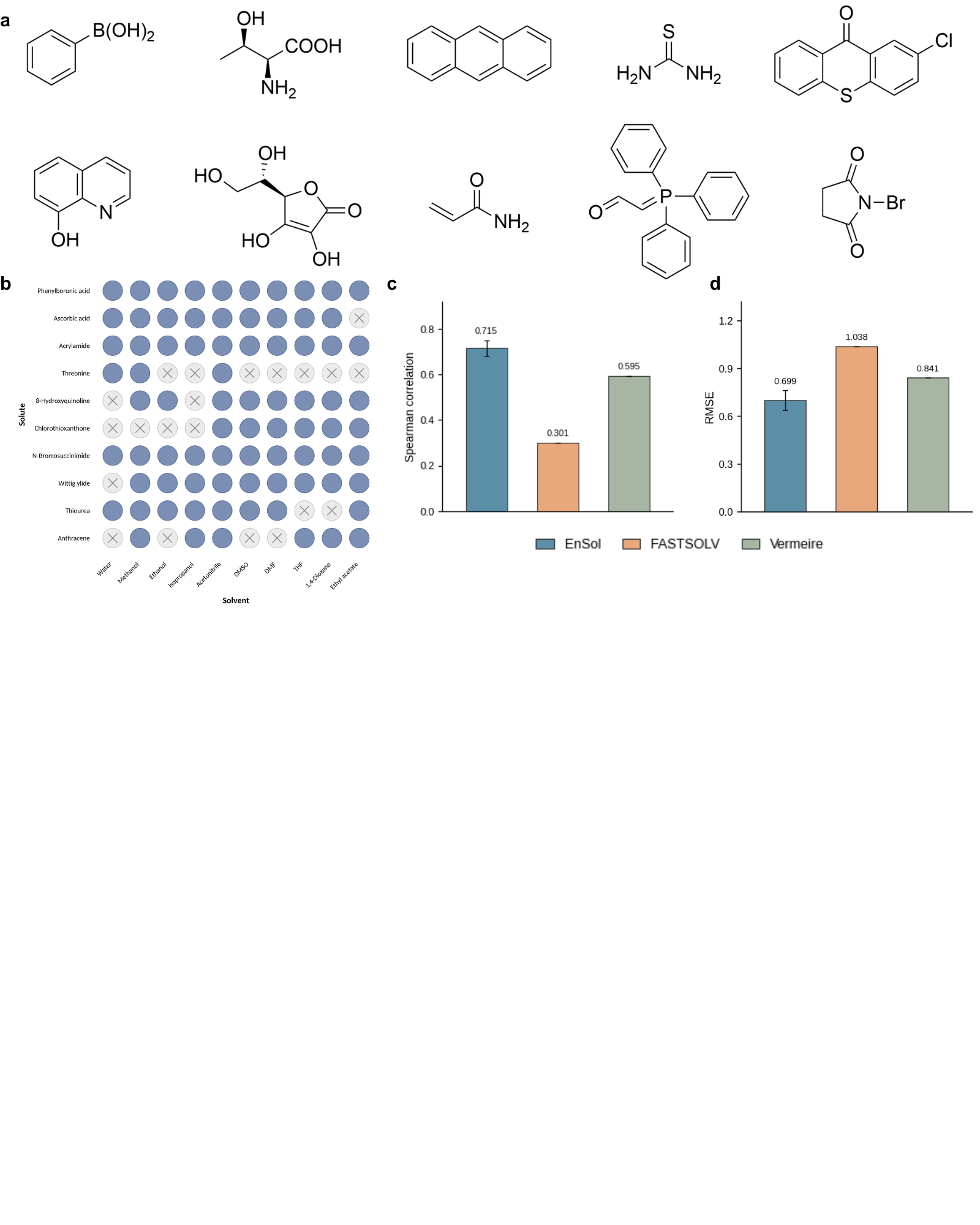}
\caption{Experimental validation set design and model performance. a) \textbf{Structural diversity of the 10 solutes used for experimental validation, spanning a range of functional groups, sizes, and scaffolds.} \textbf{b}\textbf{)} The validation set spans 10 solutes and 10 solvents, all evaluated at 298.15 K, filled circles mark the solute--solvent pairs included in the set and crossed-out circles mark pairs that were not due to the measurements below the 1 mg mL$^{-1}$ detection limit of the visual assay. \textbf{c}\textbf{)} Model performance is compared on the experimental validation set. Values above bars are mean values. EnSol error bars are standard deviation across three random seeds (n = 3) and FASTSOLV and the Vermeire model are pretrained and deterministic across Spearman rank correlation \textbf{d}\textbf{)} Model performance is compared on the experimental validation set. Values above bars are mean values. EnSol error bars are standard deviation across three random seeds (n = 3) and FASTSOLV and the Vermeire model are pretrained and deterministic across RMSE values.}\label{fig:5}
\end{figure}

Although computational benchmarks establish comparative predictive performance, molecular solubility remains an experimentally measured property, and practical utility requires that model predictions translate into reliable solvent selection under fixed laboratory conditions. To evaluate this capability, we performed targeted experimental validation, testing whether EnSol could prioritize suitable solvents beyond in silico benchmarking across a diverse set of solutes (\textbf{Fig.~\ref{fig:5}}\textbf{a}). We experimentally measured solubility across 100 chemically diverse solute--solvent pairs, comprising 10 solutes and 10 commonly used solvents selected to reflect realistic solvent-selection scenarios (\textbf{Fig. }\textbf{S}\textbf{6, Table S4}). Out of these, 78 yielded quantitative solubility values while the remaining 22 fell below the 1 mg mL$^{-1}$ detection limit of the visual assay and are therefore excluded from the quantitative metrics reported while being retained in the deposited dataset (\textbf{Fig.~\ref{fig:5}}\textbf{b}).

Compared with the computational benchmarks, this experimental dataset provides a more controlled test of model performance, as all solubility measurements were generated under a consistent experimental protocol and within a fixed experimental setting. Under these conditions, EnSol showed stronger agreement with measured solubilities than the comparison models (\textbf{Table }\textbf{S}\textbf{5}\textbf{, Fig. }\textbf{S}\textbf{7}). EnSol achieved a Spearman correlation of 0.715 $\pm$ 0.035, compared with 0.31 for FASTSOLV and 0.595 for the Vermeire model (\textbf{Fig.~\ref{fig:5}c}). The same trend was observed for prediction error, with EnSol reaching an RMSE of 0.699 $\pm$ 0.062, compared with 1.038 for FASTSOLV and 0.841 for the Vermeire model (\textbf{Fig.~\ref{fig:5}d}). Furthermore, EnSol more reliably prioritized experimentally favorable solvents compared with the baseline model. For solutes measured across multiple solvents, EnSol recovered the experimental solvent ranking with high fidelity for the majority of compounds, achieving per-solute Spearman correlations above 0.7 for six of the ten solutes and reaching 1.00 for threonine, 0.98 for thiourea, and 0.88 for 8-hydroxyquinoline (\textbf{Fig. }\textbf{S}\textbf{8}). However, this performance was not uniform across all compounds, and some solutes remained challenging for the model. Performance was weaker for a small number of solutes, most notably the phosphorane (Wittig ylide) and chlorothioxanthone. This is likely because both compounds are structurally unusual relative to typical solubility training data, with the phosphorane in particular representing a reactive organophosphorus chemistry that is poorly represented in existing datasets, leading to less reliable extrapolation. Such cases highlight an important limitation of the current model and the need for broader prospective validation across more diverse chemistries and experimental conditions.

\section{Conclusions}\label{sec:conclusions}

Solubility is one of the fundamental properties of molecules which directly affects how compounds behave under specific solvent and temperature conditions. In this work, we developed EnSol, an environment-aware probabilistic framework for molecular solubility prediction, by jointly modeling solute structure, solvent structure, and temperature-dependent environmental context. Through systematic computational evaluation on two independent external benchmarks, we showed that EnSol improves solubility ranking on the SolProp benchmark and matches the strongest available baseline under greater distribution shift. Furthermore, experimental validation across chemically diverse solutes and candidate solvents confirmed that EnSol predictions translate into practical solvent-ranking capability, to reduce experimental screening burden and improve the efficiency of early-stage molecular and process development.

We anticipate that EnSol will serve as a useful platform for data-driven solubility prediction tool in pharmaceutical development, synthetic chemistry, formulation design, and sustainable process engineering. By predicting solubility distributions, EnSol enables users to prioritize solvent candidates for experimental testing, thereby reducing screening burden, material consumption, and iteration time. Moreover, the transferability of EnSol representations provides a practical route for building expert solubility models when additional scars private or domain-specific measurements are available for a particular solute class, solvent family, or application regime.

However, despite EnSol’s capabilities, several limitations remain, including limited mechanistic interpretability, dependence on the quality and coverage of available solubility measurements, and reduced confidence in sparsely sampled regimes such as uncommon solutes, extreme temperatures, and solvent mixtures. Looking ahead, integrating EnSol with automated experimentation, active learning, and closed-loop solvent optimization could enable continuously improving solubility models that directly connect predictive uncertainty, experimental feasibility, and sustainability-aware chemical design.

\section{Methods}\label{sec:methods}

\subsection{Train and test databases}\label{sec:databases}

EnSol was trained on BigSolDB\cite{bib23}, a well-curated database of experimentally measured solubility values for organic compounds across a broad range of solvents and temperatures. Solubility values were standardized as log values of the solubility measurements. The final training dataset contained 100,570 measurements covering 1,375 solutes and 70 solvents over a temperature range of 243.15--425.77 K, following the filtering and deduplication procedures described below.

Generalization was assessed using two independent external benchmarks, SolProp\cite{bib24} covering 6,236 measurements and Leeds\cite{bib6} with 1,469 measurements. Before model training, all solutes present in either benchmark were removed from BigSolDB. The resulting benchmark evaluations therefore measure performance on compounds that were not encountered during training.

Transfer learning was evaluated using two aqueous solubility datasets with distinct sizes and chemical compositions. AqSolDB\cite{bib26} which included 9,982 compounds compiled from nine curated sources and spans a broad, heterogeneous region of drug-like and industrial chemical space. MoleculeNet ESOL\cite{bib12} contains 1,128 compounds and represents a smaller, chemically narrower and more uniformly curated benchmark. To prevent overlap with the pretraining data, compounds present in the aqueous subset of BigSolDB were removed from each target dataset using InChIKey matching. This approach captures equivalent molecular representations that may differ in their raw SMILES encoding. Deduplication removed 232 compounds from AqSolDB, leaving 9,748, and 46 compounds from ESOL, leaving 1,082. To examine how transfer performance depends on the amount of target data, fixed random subsets of 1,000 compounds were also sampled from the deduplicated AqSolDB dataset.

\subsection{Model architecture}\label{sec:architecture}

EnSol is an environment-aware molecular solubility prediction framework that jointly encodes solute and solvent molecules and explicitly conditions predictions on temperature. The model consists of graph-based molecular encoders, interaction-aware coupling between solute and solvent representations, continuous temperature conditioning via feature-wise modulation, and a probabilistic regression head for solubility prediction.

\subsubsection{Solute and solvent molecular encoders}\label{sec:encoders}

Both solute and solvent molecules are represented as molecular graphs $G=(V,E)$, where nodes correspond to atoms and are featurized using 12-dimensional atom features encoding atomic number, electronegativity, van der Waals radius, formal charge, aromaticity, hydrogen count, valence, hydrogen donor/acceptor status, and SP/SP\textsuperscript{2}/SP\textsuperscript{3} hybridization and edges correspond to their intramolecular interactions using a 6-dimensional encoding bond type (single, double, triple, aromatic), conjugation, and ring membership.\cite{bib28,bib29} Each molecule is encoded using an AttentiveFP\cite{bib26} graph neural network, performing iterative message passing to update atom-level embeddings while learning attention weights over neighboring atoms.\cite{bib30} Given initial node features $\mathbf{x}_{v}$, the AttentiveFP encoder produces a set of node embeddings ${h}_{v}$ and a graph-level embedding ${h}_{solute}$ or ${h}_{solvent}$ via pooling. The solute is encoded by a deeper AttentiveFP network (2 layers, 2 timesteps, 256 dimension) to capture more complex intramolecular interactions, while the solvent is encoded by a shallower network (1 layer, 1 timestep, 256 dimension), reflecting the asymmetric role of solvent in determining solubility context.

\subsubsection{Solute--solvent interaction modeling via cross-attention}\label{sec:cross-attention}

To model interaction-relevant context between solute and solvent molecules, EnSol employs a 4-head cross-attention to couple their learned representations.\cite{bib31} Given solute node embeddings ${H}_{s}\in\mathbb{R}^{{N}_{s}\times256}$ and solvent node embeddings ${H}_{v}\in\mathbb{R}^{{N}_{v}\times256}$, cross-attention computes context-dependent representations by allowing solute features to attend to solvent features and vice versa. Attention scores are computed as:

\begin{equation*}
\operatorname{Attention}(Q,K,V)=\operatorname{softmax}\!\left(\frac{QK^{T}}{\sqrt{d_k}}\right)V
\end{equation*}

where ${Q}_{s}={H}_{s}{W}_{Q}$ and ${K}_{v}={H}_{v}{W}_{K}$. The resulting attended representations capture interaction-relevant features conditioned on the specific solute--solvent pair. These node-level representations are pooled to produce context-aware graph embeddings.

\subsubsection{Temperature encoding and continuous conditioning}\label{sec:temperature}

Temperature is incorporated as an explicit continuous conditioning variable. Raw temperature values $T$ are first standardized using training-set statistics

\begin{equation*}
\hat{T}=\frac{T-\mu_T}{\sigma_T}
\end{equation*}

The standardized temperature is expanded using a 10-center radial basis function (RBF) expansion,

\begin{equation*}
\phi_i(T)=\exp\!\left[-\gamma_{\mathrm{rbf}}\left(\hat{T}-c_i\right)^2\right],\qquad i=1,\ldots,10
\end{equation*}

where the centers ${c}_{1},\ldots,{c}_{10}$ are placed at the 0.05--0.95 quantiles of the standardized training-temperature distribution, rather than on a fixed symmetric grid, so that RBF resolution matches the density of the training data. The bandwidth is set adaptively from the resulting center spacing as

\begin{equation*}
\gamma_{\mathrm{rbf}}=\frac{1}{2\left[\operatorname{median}\!\left(\operatorname{diff}(\operatorname{sort}(c))\right)\right]^2}
\end{equation*}

As RBF features decay to zero outside the region spanned by the centers, temperatures beyond the training range would otherwise be indistinguishable. We therefore create two monotonic extrapolation features that grow without saturating, preserving rank order into the tails

\begin{equation*}
\operatorname{overflow}(\hat{T})=\log\!\left[1+\operatorname{ReLU}\!\left(\hat{T}-\max_i(c_i)\right)\right]
\end{equation*}

\begin{equation*}
\operatorname{underflow}(\hat{T})=\log\!\left[1+\operatorname{ReLU}\!\left(\min_i(c_i)-\hat{T}\right)\right]
\end{equation*}

By appending the overflow and underflow features we end up with the temperature feature vector of

\begin{equation*}
\varphi(T)=\left[\phi_1(T),\ldots,\phi_{10}(T),\operatorname{overflow}(\hat{T}),\operatorname{underflow}(\hat{T})\right]\in\mathbb{R}^{12}
\end{equation*}

\subsubsection{Environment embedding via feature-wise linear modulation}\label{sec:film}

The temperature feature vector is projected through a two-layer perceptron that emits both scale and shift parameters for feature-wise linear modulation (FiLM):

\begin{equation*}
[\gamma_{\mathrm{raw}},\beta]=W_2\operatorname{ReLU}\!\left(W_1\varphi(T)\right)\in\mathbb{R}^{2d}
\end{equation*}

\begin{equation*}
\gamma=1+\tanh(\gamma_{\mathrm{raw}})
\end{equation*}

Given a solvent graph-level embedding ${h}_{solvent}\in\mathbb{R}^{d}$, the environment embedding is computed as

\begin{equation*}
h_{\mathrm{env}}=h_{\mathrm{solvent}}\odot\gamma+\beta
\end{equation*}

where $\odot$ denotes element-wise multiplication. Parameterizing the scale as $\gamma=1+\tanh(\gamma_{\mathrm{raw}})$ centers the transformation on the identity at initialization and constrains $\gamma\in(0,2)$, so that temperature can both attenuate and amplify individual dimensions of the solvent embedding, while the additive shift $\beta$ allows temperature to translate the representation. The resulting solvent--temperature context vector ${h}_{env}$ captures continuous thermodynamic modulation of solvent behavior. This full affine formulation replaces a scale-only variant used in preliminary experiments, in which the temperature embedding was passed through a sigmoid and applied multiplicatively; because that gate was restricted to $(0,1)$ it could only attenuate the solvent embedding and could not represent amplification or translation.

\subsubsection{Probabilistic solubility prediction via mixture density network}\label{sec:mdn}

The final solute embedding and environment embedding are concatenated and passed to a MDN implemented as a multilayer perceptron. The MDN predicts parameters of a mixture of $K=3$ Gaussian components for solubility $y$

\begin{equation*}
p(y\mid \mathbf{x})=\sum_{k=1}^{K}\pi_k(\mathbf{x})\,\mathcal{N}\!\left(y\mid\mu_k(\mathbf{x}),\sigma_k^2(\mathbf{x})\right)
\end{equation*}

where ${\pi}_{k}$ are mixture weights satisfying $\sum_{k} {\pi}_{k}=1$, and ${\mu}_{k}$ and ${\sigma}_{k}^{2}$ denote component means and variances and N is the number of training samples. The model is trained by minimizing the negative log-likelihood

\begin{equation*}
\mathcal{L}=-\sum_{n=1}^{N}\log\!\left[\sum_{k=1}^{K}\pi_k(\mathbf{x}_n)\,\mathcal{N}\!\left(y_n\mid\mu_k(\mathbf{x}_n),\sigma_k^2(\mathbf{x}_n)\right)\right]
\end{equation*}

The predictive mean is given by

\begin{equation*}
\hat{y}=\sum_{k=1}^{K}\pi_k\mu_k
\end{equation*}

and the predictive variance decomposes into aleatoric and mixture-induced uncertainty

\begin{equation*}
\operatorname{Var}(y)=\sum_{k=1}^{K}\pi_k\sigma_k^2+\sum_{k=1}^{K}\pi_k(\mu_k-\hat{y})^2
\end{equation*}

This probabilistic formulation allows EnSol to represent heteroscedastic, non-Gaussian predictive distributions and to report predictive uncertainty alongside point predictions.

\subsection{Model training and optimization}\label{sec:training}

Models were trained using the Adam optimizer with a learning rate of $1\times10^{-4}$. Training used a batch size of 32 for a maximum of 10 epochs, with the checkpoint achieving the best validation Spearman correlation retained as the final model. Training was performed on a single NVIDIA A100-SXM4-80GB GPU using PyTorch 2.7.1 and PyTorch Geometric 2.6.0. Data were partitioned at the solute level (grouped by unique solute SMILES, so no solute appeared in more than one fold) into 90:10 train--validation splits. All experiments were run with three random seeds, each controlling both model initialization and split composition, so that reported variability reflects both sources.

\subsection{Experimental workflows}\label{sec:workflows}

Solubilities of 10 molecules in 10 solvents were chosen to verify the model. Chemicals and solvents were purchased from Sigma Aldrich, Ambeed, Chemscene, Thermo Fisher Scientific, Oakwood Chemical and were used without further purification. Solubility measurements were performed by visual inspection of complete dissolution. Compounds exhibiting solubilities <1 mg mL$^{-1}$ were considered fully insoluble. For compounds with intermediate solubilities (1--10 mg mL$^{-1}$), solvent was titrated into 10 mg of substrate until a clear solution was obtained. For compounds with solubilities >10 mg mL$^{-1}$, substrate was added incrementally to 1 mL of solvent until the saturation point was reached, as indicated by the presence of persistent undissolved solid.

\subsection{Code Availability}\label{sec:code}

The source code and data for training EnSol is available at \url{https://github.com/thaonguyen217/EnSol}.

\backmatter

\section*{Acknowledgements}

This work was supported by the U.S. National Science Foundation (NSF) (CHE-2505932 (H.J. and H.Z.) and DBI-2400058 (H.Z.)). Any opinions, findings, and conclusions or recommendations expressed in this material are those of the author(s) and do not necessarily reflect those of the NSF. Our research benefitted from the computing resources at Delta, the National Center for Supercomputing Applications, enabled by allocation BIO250057 from the Advanced Cyberinfrastructure Coordination Ecosystem: Services \& Support (ACCESS) program, funded by NSF grants 2138259, 2138286, 2138307, 2137603, and 2138296.

\section*{Contributions}

H.J. and H.Z. coordinated the project. T.N and S.S jointly designed the research. T.N, S.S and Z.Z performed the research. T.N and S.S analyzed the data. T.N, S.S and Z.Z wrote the paper. All authors approved the final paper.

\section*{Declarations}

\subsection*{Competing interests}

The authors declare no competing interests.


\begin{thebibliography}{31}

\bibitem{bib1} Barrett, J. A., Yang, W., Skolnik, S. M., Belliveau, L. M. \& Patros, K. M. Discovery solubility measurement and assessment of small molecules with drug development in mind. \textit{Drug Discov. Today} \textbf{27}, 1315--1325 (2022).

\bibitem{bib2} Wu, K. \textit{et al.} Overcoming Challenges in Small-Molecule Drug Bioavailability: A Review of Key Factors and Approaches. \textit{Int. J. Mol. Sci.} \textbf{25}, 13121 (2024).

\bibitem{bib3} Tayyebi, A. \textit{et al.} Prediction of organic compound aqueous solubility using machine learning: a comparison study of descriptor-based and fingerprints-based models. \textit{J. Cheminformatics} \textbf{15}, 99 (2023).

\bibitem{bib4} Pendam, D. \textit{et al.} Advances in formulation strategies and stability considerations of amorphous solid dispersions. \textit{J. Drug Deliv. Sci. Technol.} \textbf{108}, 106922 (2025).

\bibitem{bib5} Bloomquist, C. K., Zhang, Z., Aydil, E. S. \& Modestino, M. A. Understanding the effects of multiphase flow properties in transport-limited organic electrosynthesis. \textit{Chem. Eng. J.} \textbf{528}, 172125 (2026).

\bibitem{bib6} Boobier, S., Hose, D. R. J., Blacker, A. J. \& Nguyen, B. N. Machine learning with physicochemical relationships: solubility prediction in organic solvents and water. \textit{Nat. Commun.} \textbf{11}, 5753 (2020).

\bibitem{bib7} Murdande, S. B., Pikal, M. J., Shanker, R. M. \& Bogner, R. H. Aqueous solubility of crystalline and amorphous drugs: Challenges in measurement. \textit{Pharm. Dev. Technol.} \textbf{16}, 187--200 (2011).

\bibitem{bib8} Sharma, V. \textit{et al.} Toward microfluidic continuous-flow and intelligent downstream processing of biopharmaceuticals. \textit{Lab. Chip} \textbf{24}, 2861--2882 (2024).

\bibitem{bib9} Rahimpour, E., Alvani-Alamdari, S., Acree, W. E. \& Jouyban, A. Drug Solubility Correlation Using the Jouyban--Acree Model: Effects of Concentration Units and Error Criteria. \textit{Molecules} \textbf{27}, 1998 (2022).

\bibitem{bib10} Lipinski, C. A., Lombardo, F., Dominy, B. W. \& Feeney, P. J. Experimental and computational approaches to estimate solubility and permeability in drug discovery and development settings. \textit{Adv. Drug Deliv. Rev.} \textbf{23}, 3--25 (1997).

\bibitem{bib11} Könczöl, Á. \& Dargó, G. Brief overview of solubility methods: Recent trends in equilibrium solubility measurement and predictive models. \textit{Drug Discov. Today Technol.} \textbf{27}, 3--10 (2018).

\bibitem{bib12} Delaney, John S. ESOL: estimating aqueous solubility directly from molecular structure. \textit{Journal of chemical information and computer sciences}\textit{.} \textbf{44}.3 (2004)

\bibitem{bib13} Llompart, P. \textit{et al.} Will we ever be able to accurately predict solubility? \textit{Sci. Data} \textbf{11}, 303 (2024).

\bibitem{bib14} Jouyban, A., Rahimpour, E. \& Karimzadeh, Z. A new correlative model to simulate the solubility of drugs in mono-solvent systems at various temperatures. \textit{J. Mol. Liq.} \textbf{343}, 117587 (2021).

\bibitem{bib15} Al Ibrahim, E., Morgan, N., Müller, S., Motati, S. \& Green, W. H. Accurately Predicting Solubility Curves via a Thermodynamic Cycle, Machine Learning, and Solvent Ensembles. \textit{J. Am. Chem. Soc.} \textbf{147}, 45057--45069 (2025).

\bibitem{bib16} Panapitiya, G. \textit{et al.} Evaluation of Deep Learning Architectures for Aqueous Solubility Prediction. \textit{ACS Omega} \textbf{7}, 15695--15710 (2022).

\bibitem{bib17} Francoeur, P. G. \& Koes, D. R. SolTranNet--A Machine Learning Tool for Fast Aqueous Solubility Prediction. \textit{J. Chem. Inf. Model.} \textbf{61}, 2530--2536 (2021).

\bibitem{bib18} Ghanavati, M. A., Ahmadi, S. \& Rohani, S. A machine learning approach for the prediction of aqueous solubility of pharmaceuticals: a comparative model and dataset analysis. \textit{Digit. Discov.} \textbf{3}, 2085--2104 (2024).

\bibitem{bib19} Attia, L., Burns, J. W., Doyle, P. S. \& Green, W. H. Data-driven organic solubility prediction at the limit of aleatoric uncertainty. \textit{Nat. Commun.} \textbf{16}, 7497 (2025).

\bibitem{bib20} Gao, P. \textit{et al.} Accurate predictions of drugs aqueous solubility via deep learning tools. \textit{J. Mol. Struct.} \textbf{1249}, 131562 (2022).

\bibitem{bib21} Perez, E., Strub, F., de Vries, H., Dumoulin, V. \& Courville, A. FiLM: Visual Reasoning with a General Conditioning Layer. Preprint at \url{https://doi.org/10.48550/ARXIV.1709.07871} (2017).

\bibitem{bib22} Bishop, C. M. \textit{Mixture Density Networks}. (1994).

\bibitem{bib23} Krasnov, L. \textit{et al.} BigSolDB 2.0, dataset of solubility values for organic compounds in different solvents at various temperatures. \textit{Sci. Data} \textbf{12}, 1236 (2025).

\bibitem{bib24} Vermeire, F. H., Chung, Y. \& Green, W. H. Predicting Solubility Limits of Organic Solutes for a Wide Range of Solvents and Temperatures. \textit{J. Am. Chem. Soc.} \textbf{144}, 10785--10797 (2022).

\bibitem{bib25} Ottoboni, S. \textit{et al.} A Novel Integrated Workflow for Isolation Solvent Selection Using Prediction and Modeling. \textit{Org. Process Res. Dev.} \textbf{25}, 1143--1159 (2021).

\bibitem{bib26} Sorkun, M.C., Khetan, A. \& Er, S. AqSolDB, a curated reference set of aqueous solubility and 2D descriptors for a diverse set of compounds. \textit{Sci Data} \textbf{6}, 143 (2019).

\bibitem{bib27} Wang, Z. \textit{et al.} Advanced graph and sequence neural networks for molecular property prediction and drug discovery. \textit{Bioinformatics} \textbf{38}, 2579--2586 (2022).

\bibitem{bib28} Zhao, B., Xu, W., Guan, J. \& Zhou, S. Molecular property prediction based on graph structure learning. \textit{Bioinformatics} \textbf{40}, btae304 (2024).

\bibitem{bib29} Fang, X. \textit{et al.} Geometry-enhanced molecular representation learning for property prediction. \textit{Nat. Mach. Intell.} \textbf{4}, 127--134 (2022).

\bibitem{bib30} Xiong, Z. \textit{et al.} Pushing the Boundaries of Molecular Representation for Drug Discovery with the Graph Attention Mechanism. \textit{J. Med. Chem.} \textbf{63}, 8749--8760 (2020).

\bibitem{bib31} Vaswani, A. \textit{et al.} Attention Is All You Need. Preprint at \url{https://doi.org/10.48550/ARXIV.1706.03762} (2017).

\end{thebibliography}
\end{document}